\documentclass{article}

\usepackage{ifpdf}
\usepackage{textcomp}

\usepackage{microtype}
\usepackage{graphicx}
\usepackage{subfigure}
\usepackage{booktabs} % for professional tables

\usepackage[title]{appendix}
\usepackage{float}

\usepackage[hyphens]{url}
\usepackage[breaklinks=true]{hyperref}
\usepackage{cleveref}

\usepackage[accepted]{icml2020}

\icmltitlerunning{Data-dependent Pruning to find the Winning Lottery Ticket}
\begin{document}

\twocolumn[
\icmltitle{Data-dependent Pruning to find the Winning Lottery Ticket}
% \icmltitle{In Search for Winning Lottery Tickets using Data Dependent Pruning}



\icmlsetsymbol{equal}{*}

\begin{icmlauthorlist}
\icmlauthor{D\'{a}niel L\'{e}vai}{elte}
\icmlauthor{Zsolt Zombori}{renyi,elte}
\end{icmlauthorlist}

\icmlaffiliation{renyi}{Alfr\'{e}d R\'{e}nyi Institute of Mathematics, Budapest, Hungary}
\icmlaffiliation{elte}{E{\"o}tv{\"o}s Lor\'{a}nd University, Budapest, Hungary}

\icmlcorrespondingauthor{D\'{a}niel L\'{e}vai}{levai753@gmail.com}
\icmlcorrespondingauthor{Zsolt Zombori}{zombori@renyi.hu}

\icmlkeywords{Lottery Ticket Hypothesis, Pruning, Deep Learning}

\vskip 0.3in
]



\printAffiliationsAndNotice{}  % leave blank if no need to mention equal contribution

\begin{abstract}
The Lottery Ticket Hypothesis postulates that a freshly initialized
neural network contains a small subnetwork that can be trained in
isolation to achieve similar performance as the full network. Our
paper examines several alternatives to search for such subnetworks. We
conclude that incorporating a data dependent component into the
pruning criterion in the form of the gradient of the training loss --
as done in the SNIP method -- consistently improves the performance of
existing pruning algorithms.
\end{abstract}

\section{Introduction and Related Work}
\label{introduction}

Neural network pruning techniques demonstrate that the function
learned by neural networks can often be represented with significantly
less parameters (even by less than 10\%), without compromising
performance, by selecting a small subnetwork of the initial model.

The Lottery Ticket Hypothesis, introduced in \citet{lth} postulates
that these successful subnetworks are determined not only by their
connection structure, but their initial weights as well. Once the
network is initialised, some subnetworks receive a \emph{winning ticket},
i.e., they can be trained in isolation and achieve similar performance
as if the whole network as trained.

Winning tickets have been successfully identified in a variety of
learning scenarios, e.g. \citet{lth}, \citet{lth_scale},
\citet{sparseTransferLearning}. \citet{lth_transfer} even showed that
to some extent winning tickets can be transferred across different
tasks and optimizers.

\citet{lth}, perform pruning based on the \emph{magnitude of the weights}
of a \emph{trained} network, using one or more iterations of training,
pruning and reversing to initial weights. We refer to this algorithm
as \emph{LTH}. Several alternative pruning criteria based on the
magnitude of trained and initial weights are explored in
\citet{lth_deconstructing}.

\citet{snip} introduced two fundamental differences to network
pruning. First, pruning is performed on an \emph{initialised}, but untrained
network. Second, the pruning criterion incorporates the \emph{gradient of the
  loss function with respect to the weights}. We call this algorithm as \emph{SNIP}.

Our paper argues that the more complex pruning criterion of SNIP can
be successfully combined with the more complex training-based setup of
LTH to yield a pruning algorithm that is superior to both. We argue
that the key benefit of the SNIP criterion is that it is data
dependent.

\section{Pruning Variants}
\label{sec:variants}

The difference between LTH and SNIP can be factored into two components,
which can be recombined into four different pruning algorithms.

The first factor is \emph{when} pruning happens:
\begin{itemize}
\item {\bf Training-based} pruning makes decisions based on trained
  weights. We use the iterative version of LTH: in each
  iteration, we 1) fully train the network, 2) delete $p\%$ of the
  weights then 3) revert the remaining weights to their initial value.
\item {\bf Initialisation-based} pruning makes decisions based on
  the untrained weights.
\end{itemize}

The second factor is the \emph{pruning criterion}:
\begin{itemize}
  \item {\bf Magnitude} pruning orders weights $w_i$ according
    to their absolute value $|w_i|$ and deletes the bottom $p\%$.
  \item {\bf Gradient-sensitive} pruning makes use of some training
    data $x_1, \dots x_n$. For each weight $w_i$, we compute the
    average absolute gradient of the loss with respect to the weight:
    $g_i = \frac{1}{n} \sum_{j=1}^n \left| \frac{\partial L(x_j,
      w_i)}{\partial w_i} \right|$. Next, we order weights based on
    $|w_i g_i|$ and delete the bottom $p\%$.
\end{itemize}

The possible combinations of these two factors:
\begin{itemize}
  \item {\bf Train-w}: Training-based, magnitude pruning: LTH.
  \item {\bf Train-wg}: Training-based, gradient-sensitive pruning: a
    novel approach that we argue works best.
  \item {\bf Init-w}: Initialisation-based, magnitude pruning: novel.
  \item {\bf Init-wg}: Initialisation-based, gradient-sensitive pruning: SNIP.
\end{itemize}

\section{Analysis}
\label{sec:analysis}

The ordering criterion of weights $w_i$ for magnitude pruning is $|w_i|$,
while for gradient-senstive pruning this is multiplied with the gradient of
the loss: $|w_i g_i|$. We aim to better understand
the role of the gradient component.

Gradient-sensitive pruning is more complex: weights can be
deleted either because they are small or because their contribution to the
final loss is small. As we shall see in \Cref{fig:vgg11-last-w} and
\Cref{fig:vgg11-last-wg}, there are indeed more small weights that
survive the pruning under the gradient-sensitive
criterion. Conversely, more large weights are deleted because their
gradients are small.

The extreme case is when a weight has zero corresponding gradient,
i.e., the weight simply does not contribute to the loss. The weights
of a ReLU neuron that never activates are examples of
this. Gradient-sensitive pruning very reasonably deletes such weights,
regardless of their magnitude.

The crucial property of gradient-sensitive pruning is that it
takes the dataset into consideration. The magnitude of a weight might
be a good data independent heuristic for assessing the usefulness of a
connection, while the gradient provides a data dependent alternative
heuristic. A good combination of the two likely yields an ordering
criterion that surpasses both. The product $|w_i g_i|$ is one such
combination.\footnote{One could envision other, more refined
combinations, which is left for future work.} We shall see in
\Cref{fig:vgg11-acc} that gradient-sensitive pruning indeed yields
higher test accuracy.

\section{Experiments}
\label{sec:experiments}

We run experiments in the openLTH framework~\citep{lth_repo}, using the
CIFAR-10~\citep{cifar10} dataset and the standard VGG-11,
VGG-16~\citep{simonyan2015vgg} and Resnet20~\citep{resnet}
architectures.  We use the default hyperparameters of the openLTH
framework: SGD optimizer with learning rate 0.1, momentum 0.1 and
weight decay 0.0001. We train the VGG networks for 60 epochs, the
Resnet20 for 80 epochs, and the learning rate drops to 0.01 at the
40\textsuperscript{th} and 60\textsuperscript{th} epochs,
respectively.

For training-based pruning, we perform 7 iterations, each removing
$50\%$ of the weights. For gradient-sensitive pruning, we compute the
average gradients using the whole training set.\footnote{Note that
  this is different from \citet{snip} which computes gradients using a
  single batch of data.}  Our charts mark the mean of 5 runs, and the
transparent regions mark the standard deviation.

\begin{figure}[htb]
	\centering
	\includegraphics[width=0.48\textwidth]{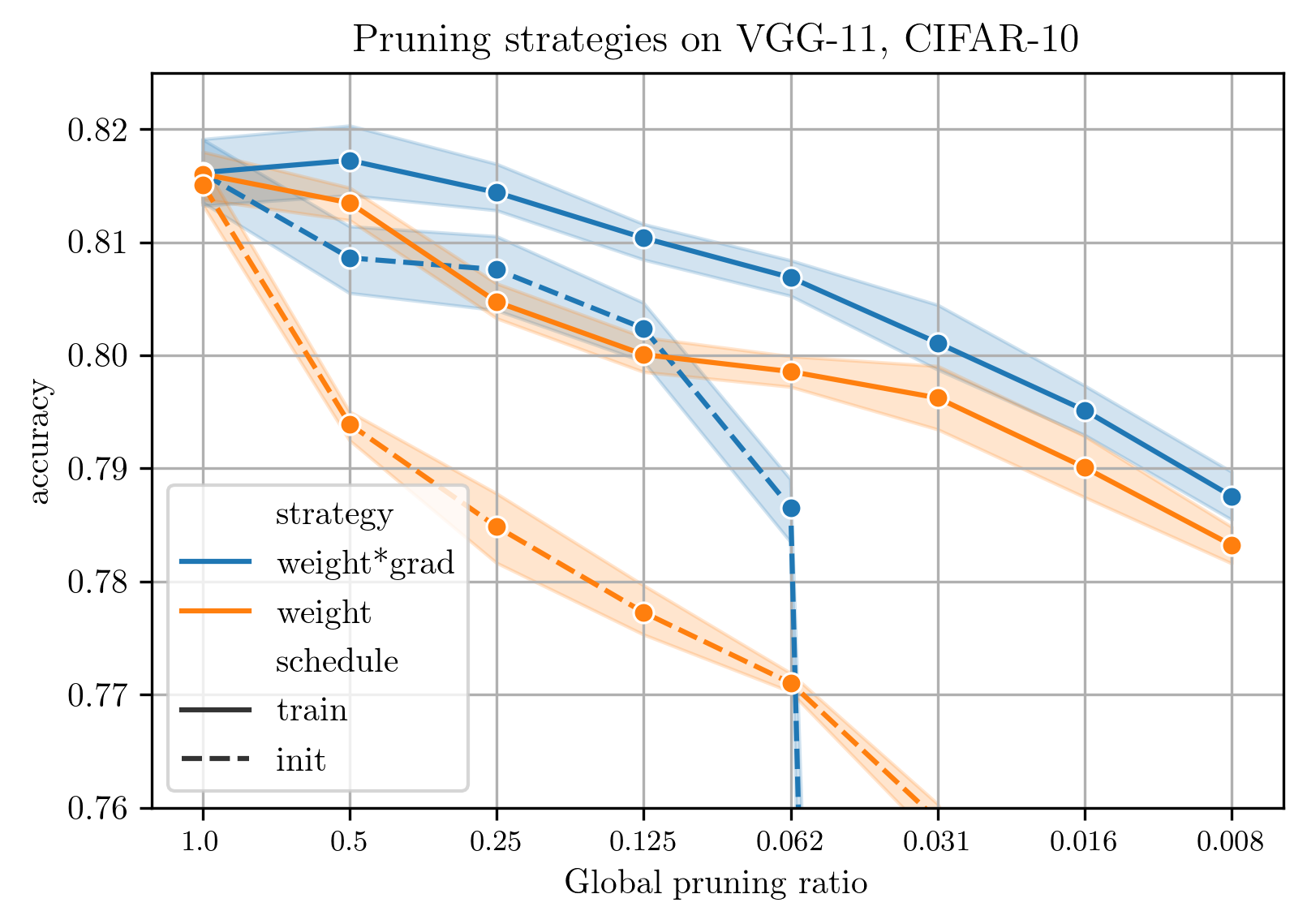}
	\caption{Accuracy of different pruning strategies on
          VGG-11. Note the logarithmic scale on the x axis.} 
	\label{fig:vgg11-acc}
\end{figure}

\begin{figure}[htb]
	\centering
	\includegraphics[width=0.48\textwidth]{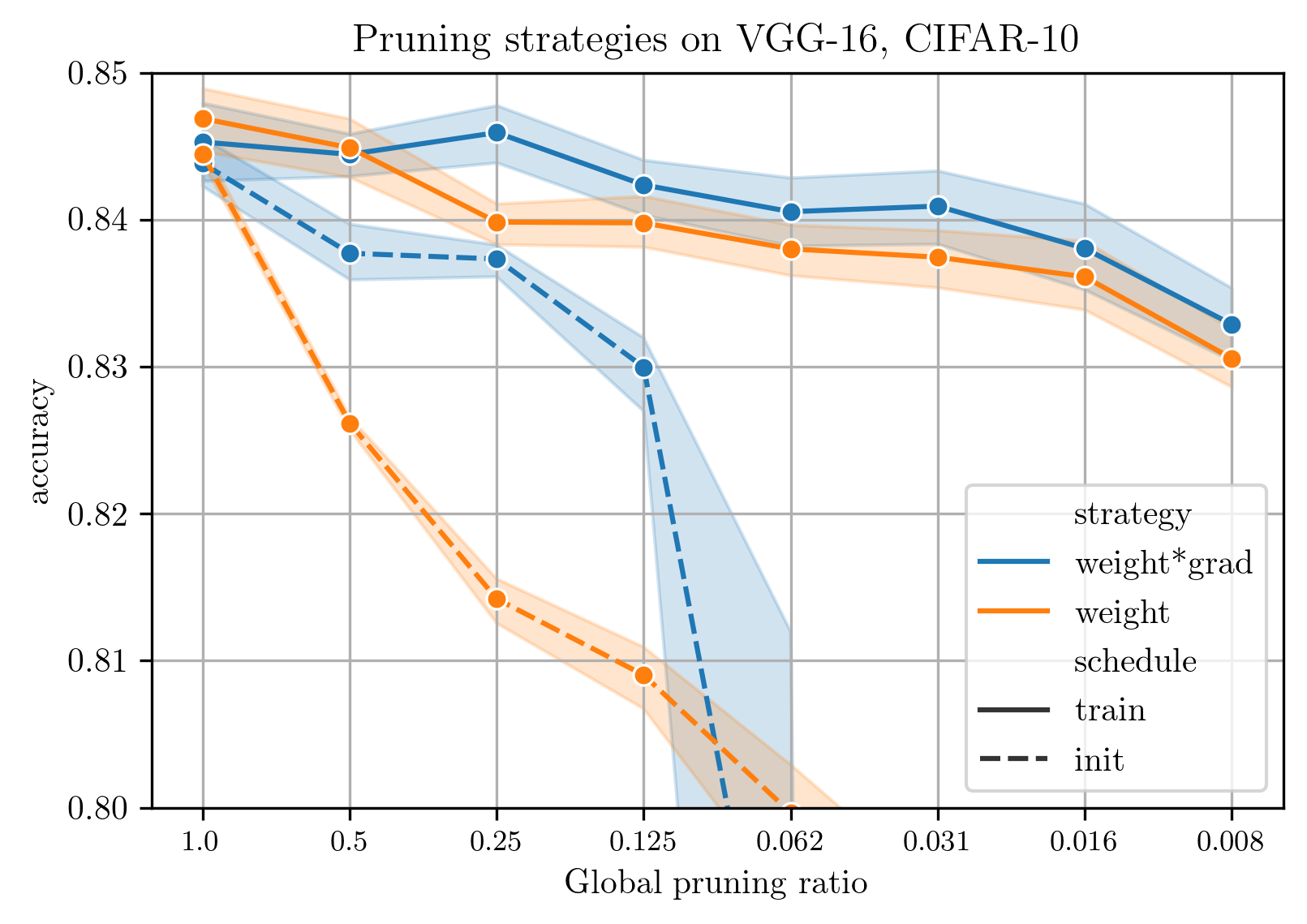}
	\caption{Accuracy of different pruning strategies on
		VGG-16. Note the logarithmic scale on the x axis.} 
	\label{fig:vgg16-acc}
\end{figure}

\Cref{fig:vgg11-acc} and \Cref{fig:vgg16-acc} show the differences in
accuracy for the four strategies introduced in
Section~\ref{sec:variants} for VGG-11 and VGG-16.  Training-based
pruning methods clearly outperform initialisation-based methods by a
large margin, and gradient-sensitive methods outperform magnitude
methods by a smaller margin.  At lower pruning levels ($50\%-6\%$),
the difference of the two pruning criteria is larger.
%but it fades away at extreme pruning levels ($6\%-0.8\%$).

\begin{figure}[!t]
	\centering
	\includegraphics[width=0.48\textwidth]{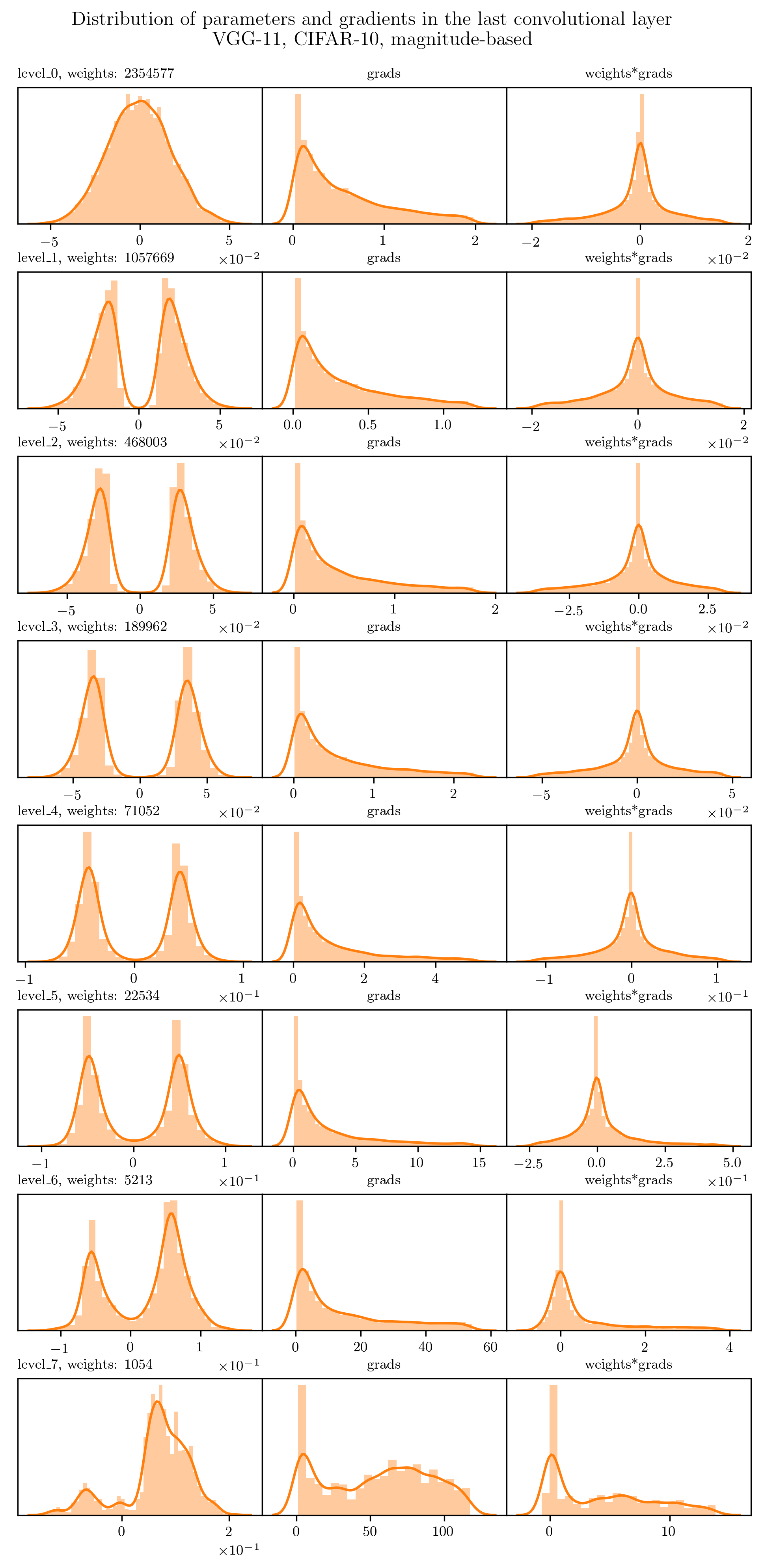}
	\caption{Weights, gradients and weights $\cdot$ gradients
          in the final conv layer, before pruning happens. Magnitude
          pruning.} 
	\label{fig:vgg11-last-w}
\vspace{-7pt}
\end{figure}

\begin{figure}[!t]
	\centering
	\includegraphics[width=0.48\textwidth]{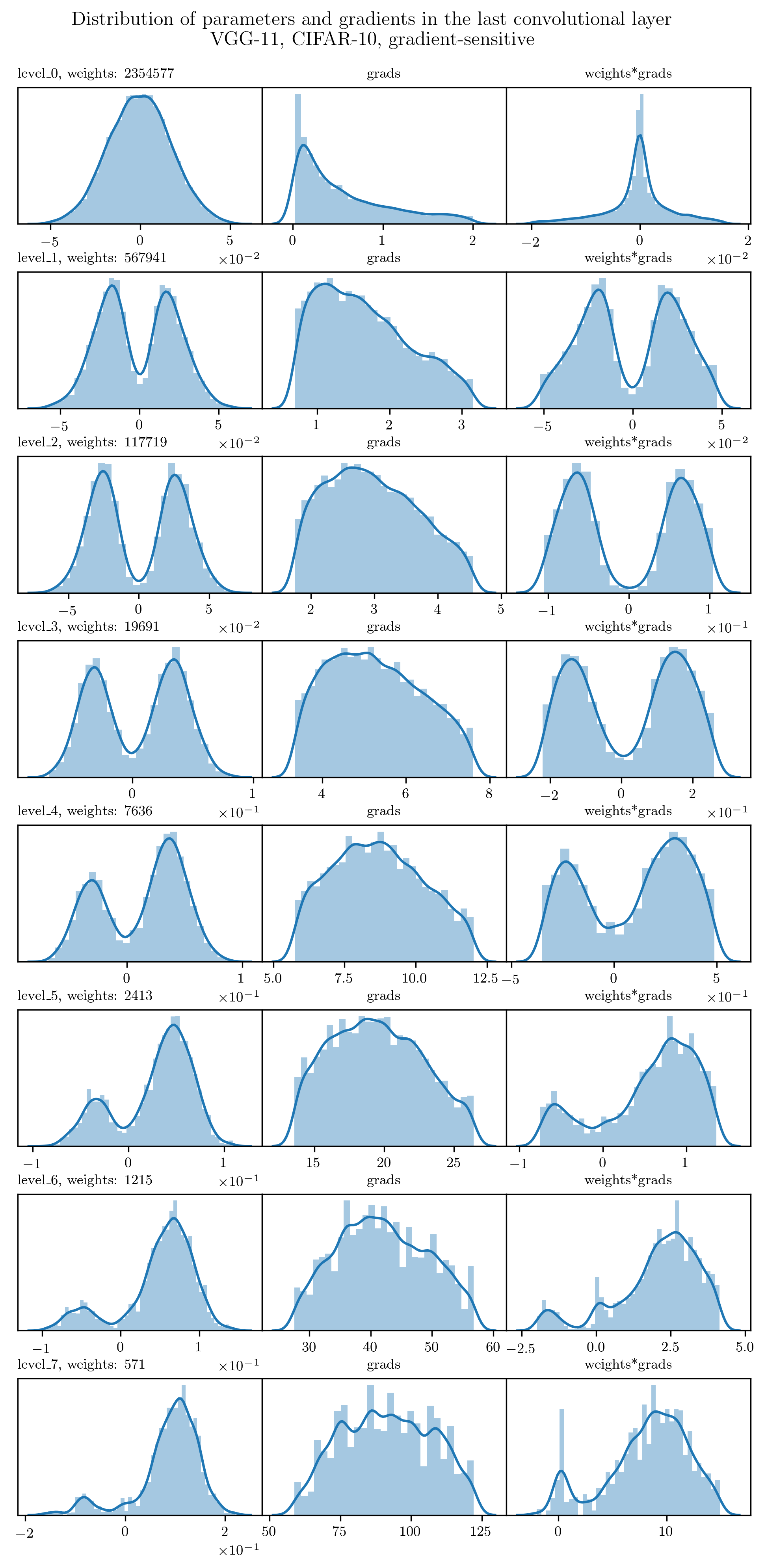}
	\caption{Weights, gradients and weights $\cdot$ gradients
          in the final conv layer before pruning happens. Gradient-sensitive pruning.} 
	\label{fig:vgg11-last-wg}
\vspace{-7pt}
\end{figure}

\begin{figure}[htb]
	\centering
	\includegraphics[width=0.48\textwidth]{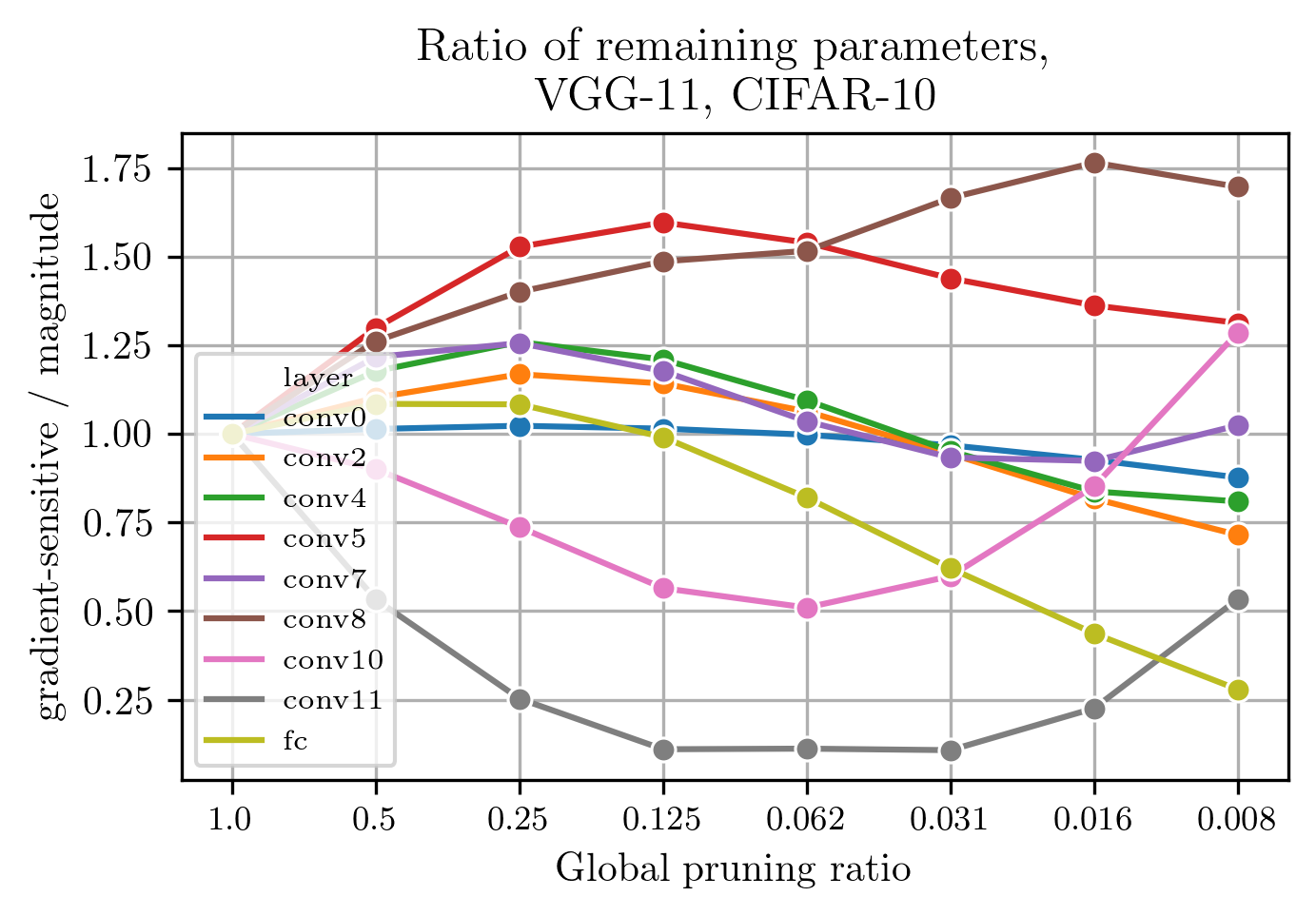}
	\caption{Number of remaining weights under gradient-sensitive
          pruning divided by the number of remaining weights under
          magnitude pruning for each layer. Training-based setup,
          VGG-11.}
	\label{fig:vgg11-ratio}
\end{figure}

\begin{figure}[htb]
	\centering
	\includegraphics[width=0.48\textwidth]{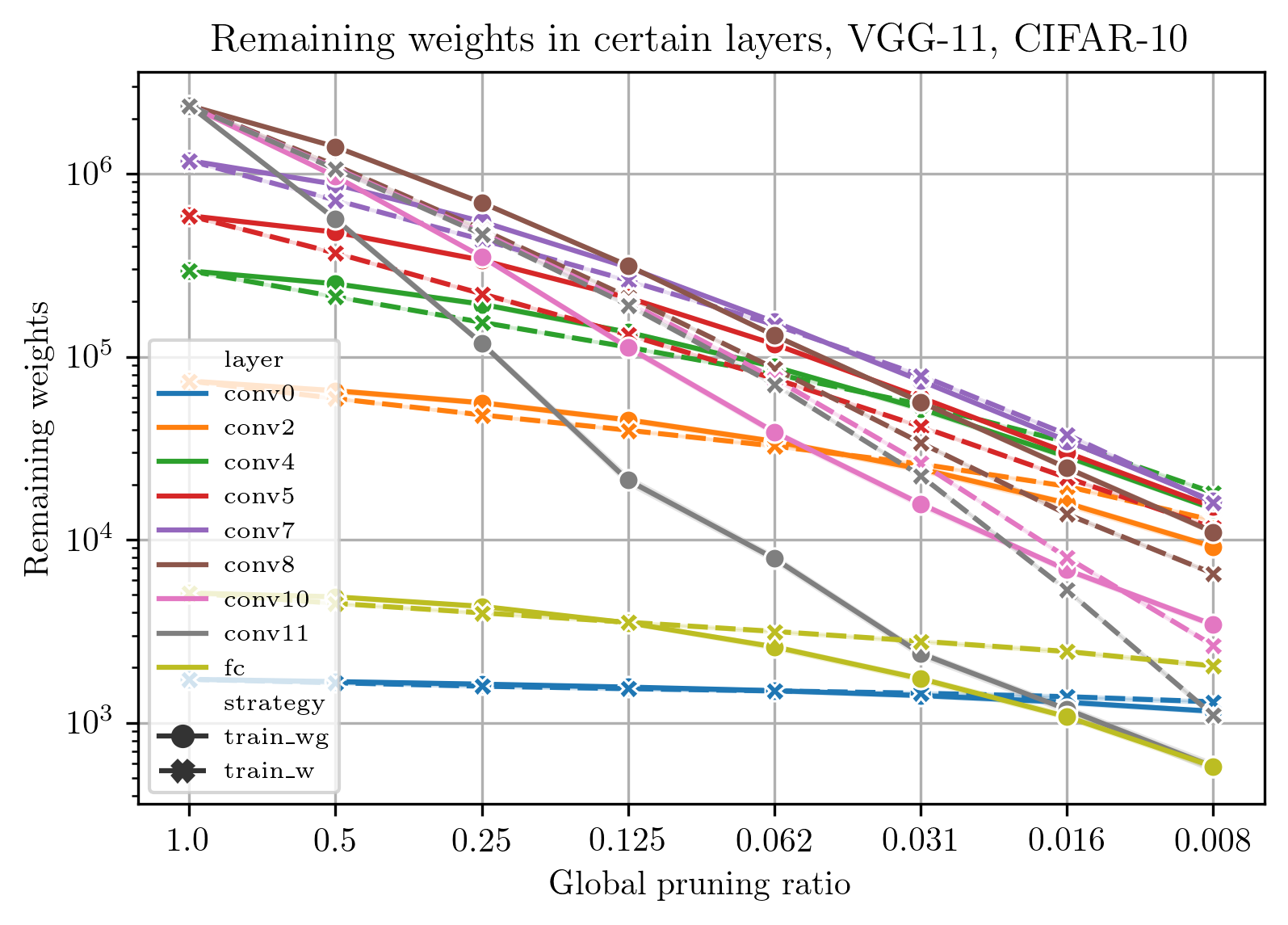}
	\caption{Number of remaining weights by layer for
          the two training
          based pruning strategies on VGG-11.}
	\label{fig:vgg11-remaining}
\end{figure}

Gradient-sensitive pruning results in higher test accuracy. To get a
deeper insight into this result, we look at layerwise differences in
the training-based setting. \Cref{fig:vgg11-ratio} shows that
gradient-sensitive pruning eliminates slightly more weights from the
first layers and significantly more from the last layers, keeping more
parameters in the middle layers. At extreme sparsity, we can see from
\Cref{fig:vgg11-remaining} that gradient-sensitive pruning keeps very
few weights from the last layers: above $99\%$ pruning, less than 1000
weights remain in the last convolutional layer, and the classifying
dense layer undergoes even more significant pruning.

Diving deeper, we focus on the last convolutional layer in the
training-based setups and examine histograms of weights, gradients and
weights times gradients for magnitude pruning
(\Cref{fig:vgg11-last-w}) and gradient-sensitive pruning
(\Cref{fig:vgg11-last-wg}).

The left columns show the distribution of weights. Notice that pruning
leaves a ``hole'' in the histogram of subsequent iterations around zero. This
means that after we remove small weights, reverting and retraining
barely yields new small weights. As a result, subsequent
iterations prune increasingly larger weights. 

% The histograms suggest
% that the weights roughly return to the same value in each
% iteration.

The hole is less accentuated for gradient-sensitive pruning, which is
expected, since some of the small weights survive pruning due to
having larger gradients. On the other hand, the holes appear for
gradient-sensitive pruning in the product of the weights and the
gradients (right side of Figure \ref{fig:vgg11-last-wg}), since this
is the pruning criterion. Note, however, that these holes are less
rigid than their magnitude based counterparts and disappear in fewer iterations. This is because
while the trained weights do not seem to be greatly affected by
pruning, their gradiens adapt more easily.

Finally, the middle columns of Figure~\ref{fig:vgg11-last-w} and
\ref{fig:vgg11-last-wg}, reveal the distribution of the average
absolute values of the gradients. At the end of the first training, we
see a large number of weights with zero gradients. Gradient-sensitive
pruning removes all of them and yields rather healthy gradient
histograms in later iterations. Magnitude-based pruning, on the other
hand, keeps these weights, even though they are completely
useless. This shows that the magnitude of a weight is not a good proxy
for the magnitude of the gradient.

\begin{figure}[htb]
	\centering
	\includegraphics[width=0.48\textwidth]{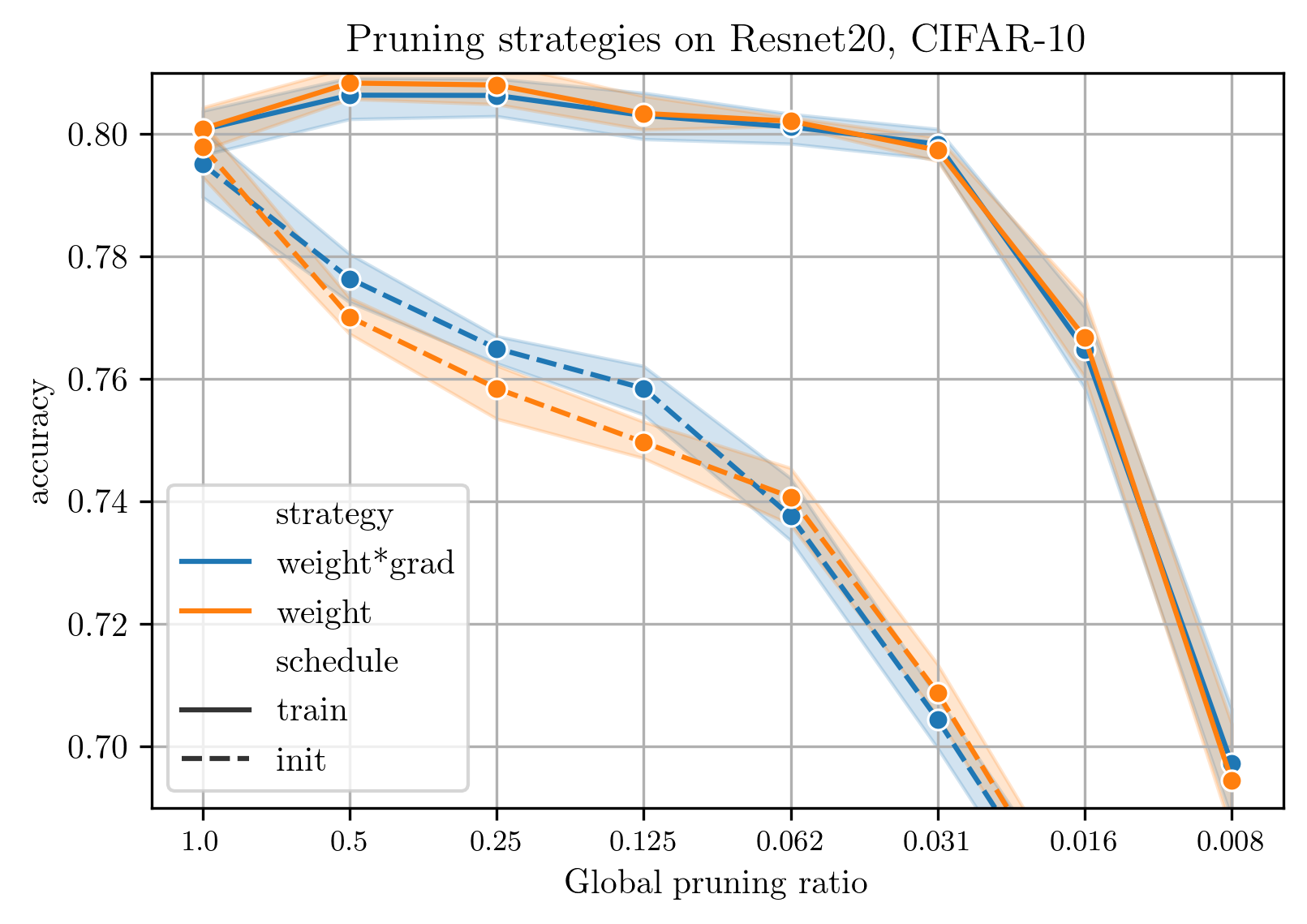}
	\caption{Accuracy of different pruning strategies on
		ResNet20. Note the logarithmic scale on the x axis.} 
	\label{fig:resnet20-acc}
\end{figure}

\begin{figure}[!ht]
	\centering
	\includegraphics[width=0.48\textwidth]{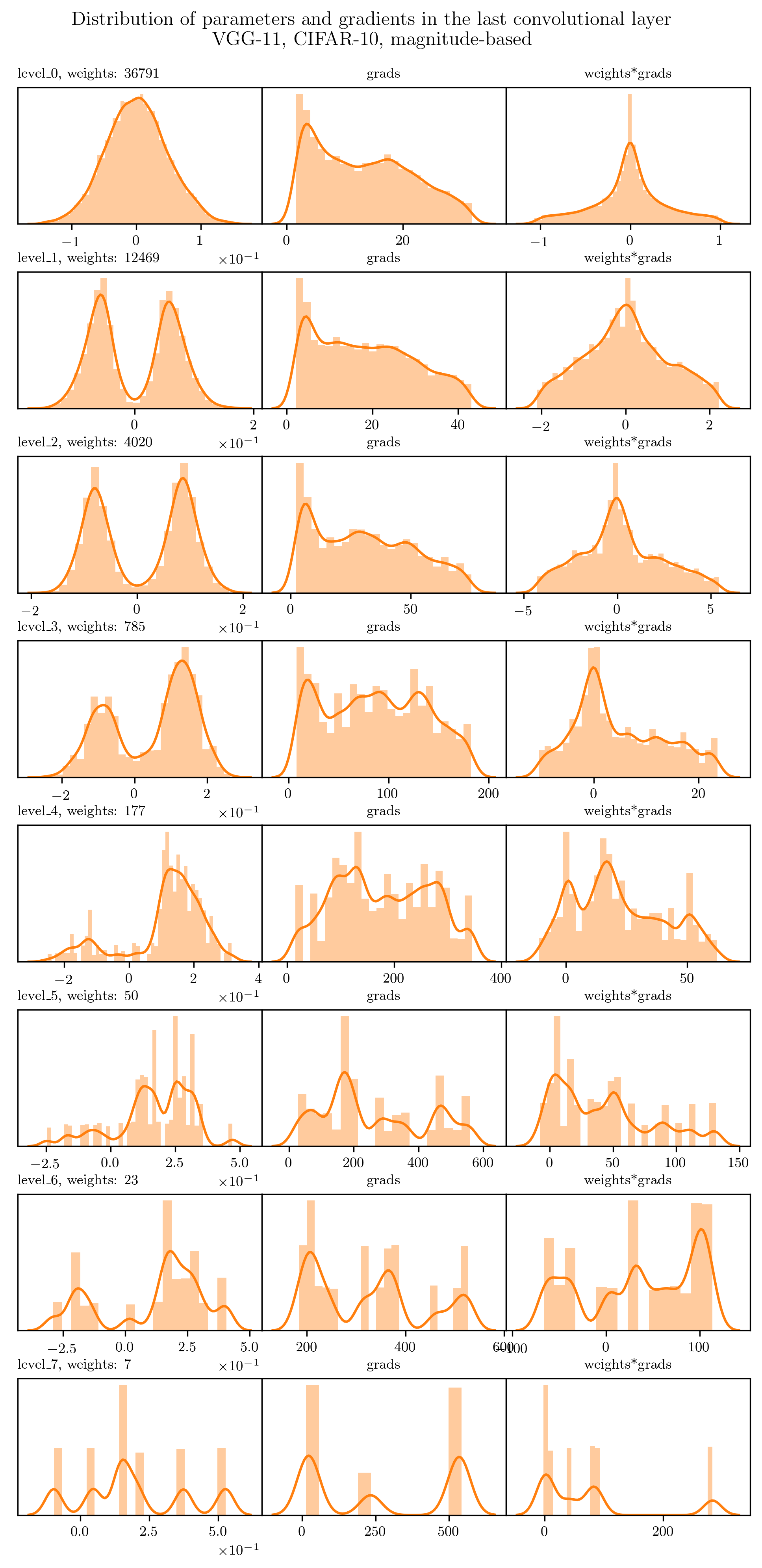}
	\caption{Histogram of weights, gradients and weights $\cdot$ gradients
		in the final network layer of ResNet20, before pruning happens. Magnitude
		pruning.} 
	\label{fig:resnet20-last-w}
\end{figure}

\begin{figure}[!ht]
	\centering
	\includegraphics[width=0.48\textwidth]{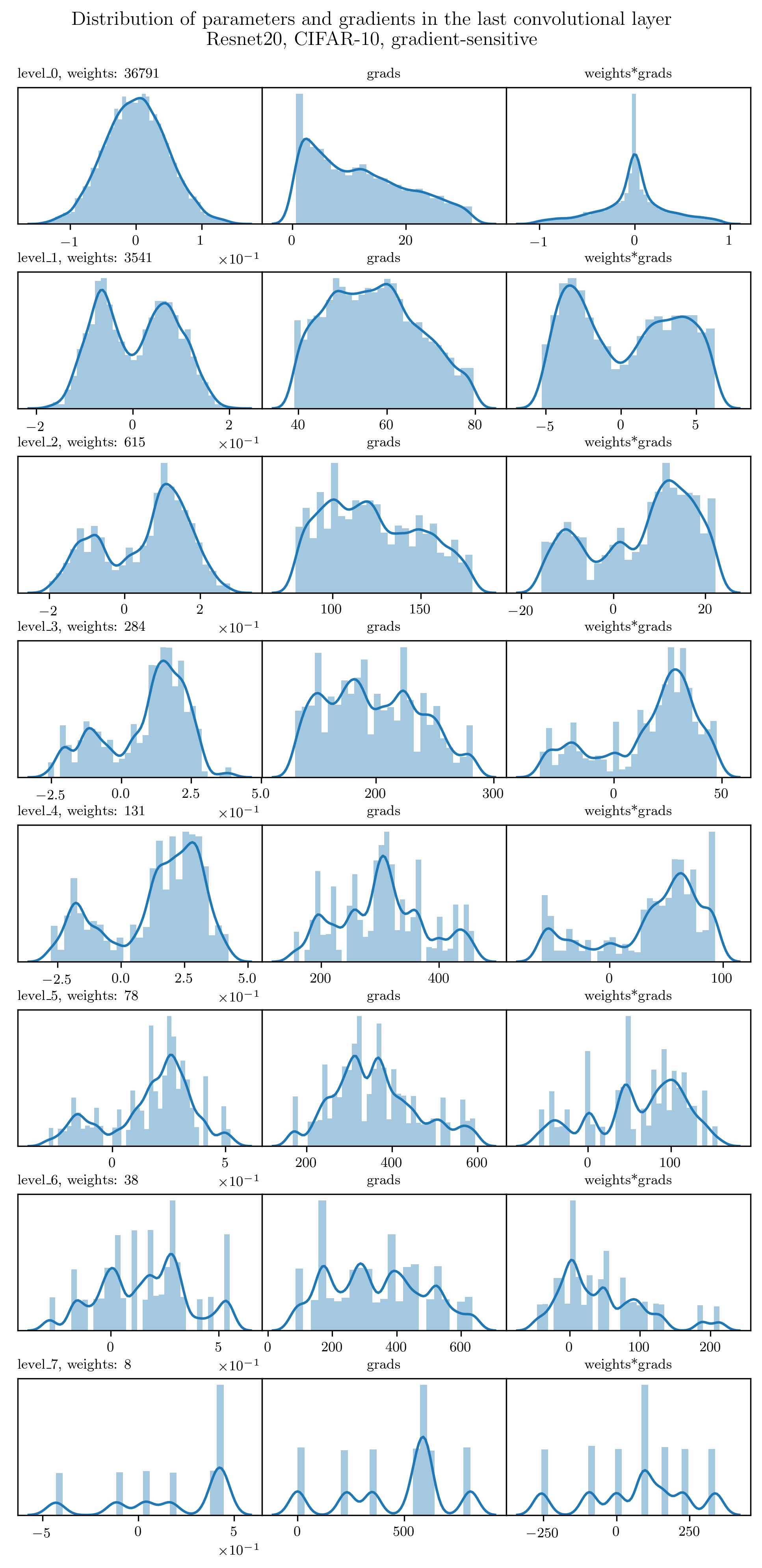}
	\caption{Histogram of weights, gradients and weights $\cdot$ gradients
		in the final network layer or ResNet20, before pruning happens. Gradient-sensitive pruning.} 
	\label{fig:resnet20-last-wg}
\end{figure}

\paragraph{ResNet20}
\Cref{fig:resnet20-acc} shows the training accuracies on
ResNet20. Interestingly, the network is much less resistant to
pruning, which we conjecture is due to its higher parameter
efficiency. Furthermore, there is no significant difference between
magnitude and gradient-sensitive pruning. The residual connections
allow for healthier gradient flow, which we can see from
\Cref{fig:resnet20-last-w} and \ref{fig:resnet20-last-wg}. There are
much fewer parameters with extremely small gradients. Deeper
understanding of the pruning behaviour of residual networks is left
for future work.

\section{Conclusion and Future Work}

Our work compares four network pruning algorithms that are the
combinations of a simple (initialisation-based) and a complex (training-based)
pruning setup on one hand and a simple (magnitude-based) and a complex
(gradient-sensitive) pruning criterion on the other. We show that
training-based approaches surpass initialization-based approaches and
that gradient-sensitive approaches surpass magnitude-based approaches.

A key takeaway message from our work is the benefit of a data
dependent pruning criterion as manifested in the gradient-sensitive
pruning scenarios. 

As future work, we note that the gradient-sensitive formula used in
\citet{snip} is one, but not necessarily the best solution. For
example, we could introduce an extra exponent parameter $\lambda$ and order by
$\left|w_i g_i^\lambda\right|$.
Another promising direction is to find a good interpolation between
initialization-based and training-based pruning. A lot of evidence
suggests that training allows for better pruning, however, it may not
be necessary to fully train the network to make good pruning
decisions.  This promises to match the performance of
training-based variants, without the large computational overhead.
Futhermore, we are interested to see if there are systematic
differences between the subnetworks selected by different pruning
strategies. \Cref{fig:vgg11-remaining} already reveals differences in
the layerwise number of remaining weights, and we are intrigued to
find more complex patterns. 

%% Another interesting direction is to try to
%% identify components larger than weights that are selected during
%% pruning. These inquiries can lead us to better understand the
%% structure of ``winning tickets'' and possibly allow for designing
%% architectures that are a priori closer to such subnetworks. Another
%% approach is to try to incorporate the search for winning tickets into
%% the optimisation procedure by explicitly rewarding a certain structure.

\section{Acknowledgments}
This work has been supported by the European Union, co-financed by the
European Social Fund (EFOP-3.6.3-VEKOP-16-2017-00002), as well as by
the Hungarian National Excellence Grant 2018-1.2.1-NKP-00008.

\bibliography{pruning}
\bibliographystyle{icml2020}

\end{document}